\documentclass[11pt]{article}

\usepackage[final]{acl}

\usepackage{times}
\usepackage{latexsym}

\usepackage[T1]{fontenc}
\usepackage[utf8]{inputenc}

\usepackage{microtype}

\usepackage{inconsolata}

\usepackage{graphicx}
\usepackage{subfigure}
\usepackage{algorithm}
\usepackage{algorithmic}

\usepackage{booktabs}
\usepackage{multirow}
\usepackage{makecell}

\usepackage{amsmath,amssymb}

\usepackage{comment}

\newcommand{\mycomment}[1]{}

\usepackage[most,skins,theorems]{tcolorbox}
\tcbset{
  aibox/.style={
      width=\linewidth,
      top=7pt,
      bottom=2pt,
      colback=blue!6!white,
      colframe=black,
      colbacktitle=black,
      enhanced,
      center,
      attach boxed title to top left={yshift=-0.1in,xshift=0.15in},
      boxed title style={boxrule=0pt,colframe=white,},
    }
}
\newtcolorbox{AIbox}[2][]{aibox,title=#2,#1}

\usepackage{enumitem}

\usepackage{listings}
\title{SCAN: Semantic Document Layout Analysis for Textual and Visual Retrieval-Augmented Generation}

\author{
  Nobuhiro Ueda\footnotemark[1],
  Yuyang Dong\thanks{Equal contribution}\thanks{Current affiliation: SB Intuitions Corp. Email:\\\texttt{yuyang.dong@sbintuitions.co.jp}},
  Krisztián Boros,
  Daiki Ito,
  Takuya Sera,
  Masafumi Oyamada \\
  NEC Corporation \\
  \texttt{\{nobuhiro-ueda,dongyuyang,krisztian-boros,ito-daiki,takuya-sera,oyamada\}@nec.com}
}

\begin{document}

\maketitle

\begin{abstract}
  With the increasing adoption of Large Language Models (LLMs) and Vision-Language Models (VLMs),
  rich document analysis technologies for applications like Retrieval-Augmented Generation (RAG)
  and visual RAG are gaining significant attention.
  Recent research indicates that using VLMs yields better RAG performance,
  but processing rich documents remains a challenge since a single page contains large amounts of information.
  In this paper, we present SCAN (\textbf{S}emanti\textbf{C} Document Layout \textbf{AN}alysis),
  a novel approach that enhances both textual and visual Retrieval-Augmented Generation (RAG) systems
  that work with visually rich documents.
  It is a VLM-friendly approach that identifies document components with appropriate semantic granularity,
  balancing context preservation with processing efficiency.
  SCAN uses a coarse-grained semantic approach that divides documents into coherent regions covering contiguous components.
  We trained the SCAN model by fine-tuning object detection models on an annotated dataset.
  Our experimental results across English and Japanese datasets demonstrate that applying SCAN improves
  end-to-end textual RAG performance by up to 9.4 points and visual RAG performance by up to 10.4 points,
  outperforming conventional approaches and even commercial document processing solutions.
\end{abstract}

\section{Introduction}
Retrieval-Augmented Generation (RAG)~\cite{lewis2021rag, gao2024ragsurvey,fan2024ragllmsurvey} technology enables Large Language Models (LLMs) to provide more accurate responses to user queries by retrieving and leveraging relevant knowledge and documents.
These knowledge sources, such as company financial reports, web pages, insurance manuals, and academic papers, often contain complex charts, tables, diagrams, and other non-textual elements, collectively referred to as rich documents.
Effectively enabling RAG systems to understand and utilize such rich, multimodal content remains a key research challenge.

In practice, RAG systems for rich documents follow two major pipeline patterns.
Textual RAG first converts documents into text (e.g., Markdown) format,
performs text retrieval, and then generates responses with an LLM.
Visual RAG, by contrast, retrieves images of pages or regions and uses a Vision-Language Model (VLM) to read the images and generate an answer directly.
Although their modalities differ, both pipelines ultimately depend on VLMs---either to convert visual regions into text or to interpret them directly---and both tend to break down when an entire page is processed at once.

\begin{figure*}[t]
  \centering
  \begin{minipage}{0.46\textwidth}
    \includegraphics[width=\textwidth]{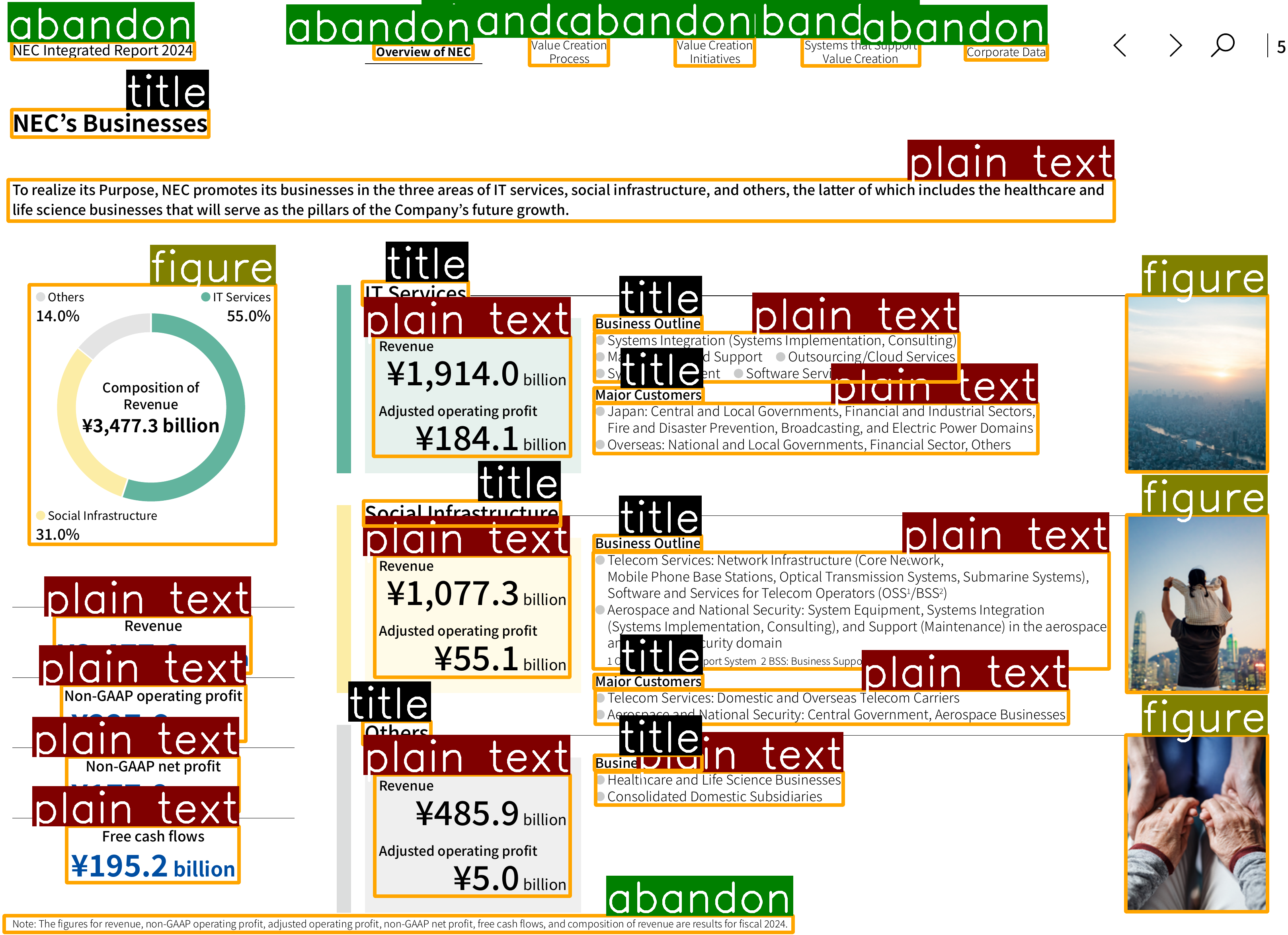}
  \end{minipage}
  \hspace{5mm}
  \begin{minipage}{0.46\textwidth}
    \includegraphics[width=\textwidth]{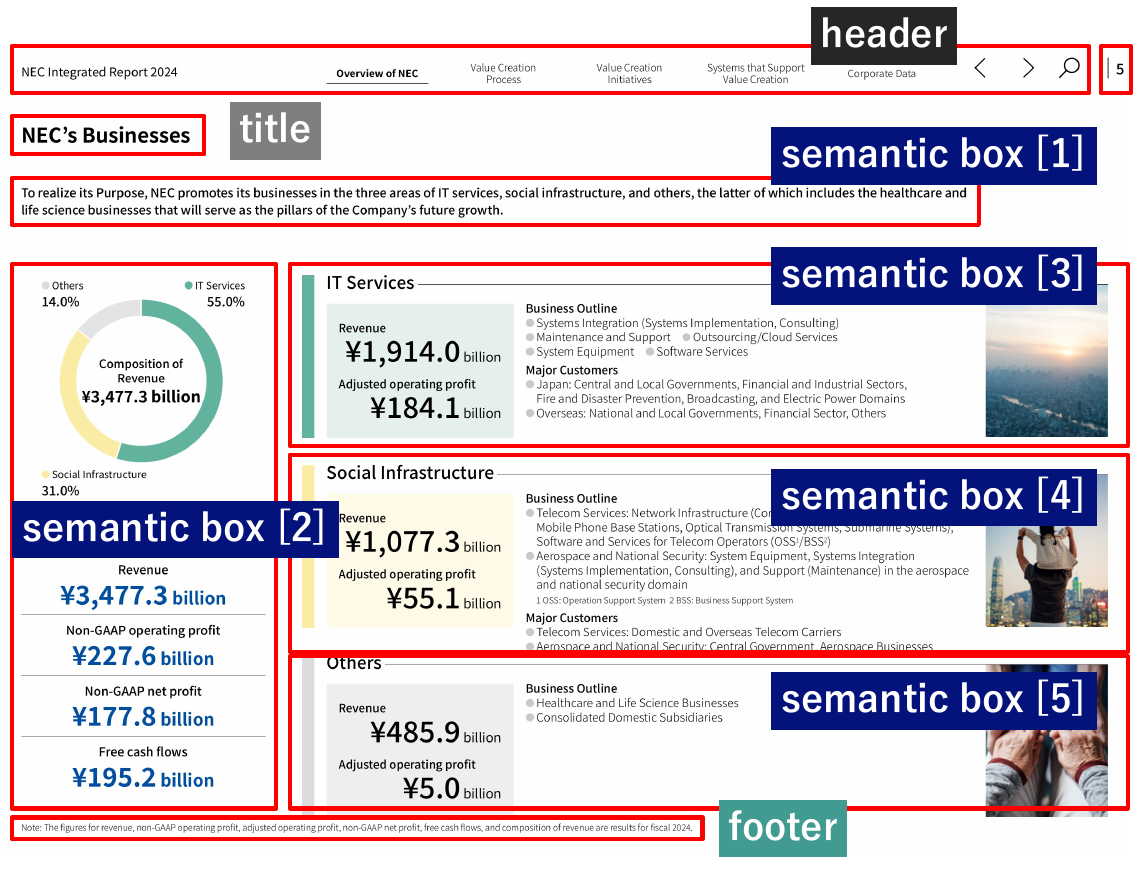}
  \end{minipage}
  \caption{Conventional fine-grained layout analysis result (left, DocLayout-YOLO) vs. our coarse-grained semantic layout analysis result (right, SCAN).}
  \label{fig:layoutvs}
\end{figure*}

\begin{figure*}[t]
  \centering
  \includegraphics[width=0.9\textwidth]{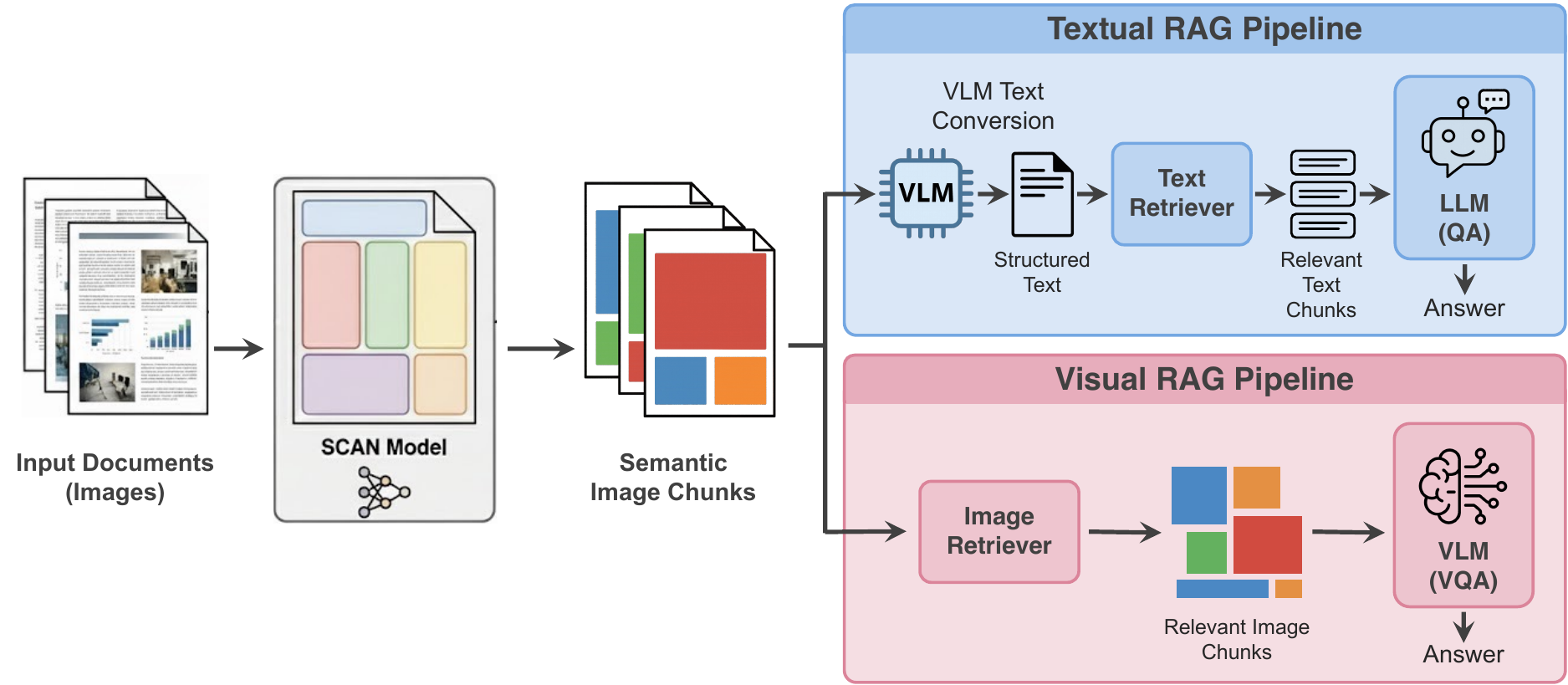}
  \caption{Overview of applying our SCAN model to current textual and visual RAG pipelines.}
  \label{fig:system-overview}
\end{figure*}



Therefore, \textbf{a common challenge is having a VLM process an entire rich document page (text conversion or VQA) at once.}
One potential solution is to further divide a document page into small regions.
Traditional document layout analysis technologies such as DocLayout-YOLO~\cite{doclayoutyolo} can achieve this objective, but they focus on fine-grained analysis, breaking down content into small components such as titles, paragraphs, tables, figures, and captions.
This approach could lose important context when processing isolated components and potentially lead to reduced RAG accuracy.
In our experiments, conventional layout analysis methods with VLM text conversion and VQA degraded RAG performance in most cases.

\subsection{Contributions}
To address these challenges, we propose SCAN, a novel approach that performs VLM-friendly semantic document layout analysis with ``coarse granularity.''
Figure~\ref{fig:layoutvs} compares the result of conventional layout analysis with that of SCAN.
SCAN can semantically divide regions into boxes that cover contiguous components related to the same topic.
For example, each of the semantic boxes [3], [4], and [5] corresponds to independent topics of \textit{IT Services}, \textit{Social Infrastructure}, and \textit{Others}.

To train a powerful SCAN model, we annotated more than 24k document pages with semantic layout labels.
The model is fine-tuned from pre-trained object detection models using this training data.
We also designed post-processing techniques for RAG applications.
Figure~\ref{fig:system-overview} gives an overview of applying our SCAN model to both textual and visual RAG pipelines.
Concretely, each page of an input document is treated as an image and decomposed into semantic chunks by our SCAN model.
For textual RAG, the images of semantic chunks are passed to a VLM that performs text conversion, and the resulting texts are then input into existing textual RAG systems.
For visual RAG, the resulting image chunks can be the retrieval targets that are directly input into existing visual RAG systems.

We evaluate SCAN's performance using three datasets featuring both English and Japanese documents.
Although SCAN is trained on Japanese data, the experiments show that it can achieve good performance on English benchmarks.
Experimental results show that in textual RAG, applying SCAN can improve end-to-end performance by 2.7--9.4 points, while in visual RAG, SCAN can enhance end-to-end performance by 5.6--10.4 points.
Moreover, although SCAN requires multiple VLM inferences rather than a single page-level inference, the total computational time is reduced because each inference uses fewer input tokens.

%
%
%

\section{Related Work}

\subsection{Document Layout Analysis}
CNN-based models such as DocLayout‐YOLO~\citep{doclayoutyolo} and Transformer-based models such as DiT~\citep{li2022dit}, LayoutLMv3~\citep{huang2022layoutlmv3}, and Beehive~\citep{docling} have been proposed for high-performance document layout analysis.
These models are trained with synthetic and human-annotated datasets~\cite{publaynet,docbank,doclaynet2022}.
Building on these models, several end-to-end document conversion systems have been developed.
Docling~\citep{docling}, Marker~\citep{marker}, and MinerU~\citep{mineru} provide a comprehensive pipeline for document layout analysis and text conversion.
In addition, production systems such as Azure Document Intelligence~\cite{adi} and LlamaParse Premium~\cite{llamaparse} are available.

\subsection{VLMs for Document Conversion}
Vision-Language Models (VLMs) have emerged as powerful tools for multimodal understanding tasks.
Open models such as the Qwen-VL series~\cite{Qwen2.5-VL,qwen3-vl} and InternVL~\cite{internvl}
have also demonstrated impressive capabilities in visual document understanding.
Moreover, smaller OCR-specialized VLMs, including GOT~\cite{got}, Nougat~\cite{nougat}, DocVLM~\cite{docvlm}, and olmOCR~\cite{olmocr}, have been developed to efficiently handle document text extraction.

\subsection{Textual and Visual RAG}
The rapid progress of LLMs has further strengthened RAG,
enabling models to inject external knowledge into responses with higher precision and coverage~\cite{lewis2021rag,gao2024ragsurvey,fan2024ragllmsurvey}.
Typical RAG pipelines in previous works first converted document images or PDFs into plain text,
and only then performed indexing and retrieval over the extracted passages~\cite{ohr}.
Recent results show that using VLMs for document text conversion yields better results than traditional OCR tools~\cite{ohr,ocrbenchv2}.
On the other hand, with the increasing availability of multimodal embeddings and VLMs, there is a growing interest in multimodal RAG systems that can directly index and retrieve images and leverage VLMs for answer generation~\cite{visrag,vdocrag,colpali}.

\section{Method}

Our goal is to design a VLM-friendly layout analysis module that divides a rich document page into a small number of semantically coherent regions with coarse granularity.
We call this task \emph{semantic document layout analysis}, and the resulting regions \emph{semantic chunks}.
In contrast to conventional layout analysis, which produces many fine-grained boxes for structural elements (titles, paragraphs, tables, figures, etc.), SCAN aims at a coarser granularity that better matches how humans understand a page and how RAG systems use it.

A semantic chunk on a page is defined as a region whose content is unified by a single subtopic.
Complex pages often contain several subtopics under one broader topic.
For example, in the page shown in Figure~\ref{fig:layoutvs}, the overall topic is ``business areas of a company,'' but there are three distinct subtopics: \textit{IT Services}, \textit{Social Infrastructure}, and \textit{Others}.
Each of these subtopics is represented by one semantic chunk.
Formally, we require that everything inside one chunk is necessary to understand that subtopic and that no important part of the subtopic lies outside the chunk.
This notion is different from \emph{structural} divisions such as paragraphs, sections, or table cells.
Structural divisions are defined by superficial document elements (e.g., line breaks or section headers) and do not necessarily align with topical boundaries.
They are particularly unreliable for floating elements such as figures and tables, or for infographic-style pages where text and graphics are freely laid out.
Therefore, we define semantic chunks directly in terms of topical coherence rather than layout structure.
In our implementation, semantic chunks are represented as rectangular bounding boxes, and we introduce two types of boxes.
\emph{Semantic boxes} correspond to content related to one subtopic as defined above; one page typically contains several semantic boxes.
\emph{Global boxes} correspond to page-level metadata such as the title, header, footer, date, and author.
Global boxes are semantically related to the whole page, whereas semantic boxes are local and mostly independent of each other.
Figure~\ref{fig:layoutvs} (right) illustrates five semantic boxes and three global boxes produced by our SCAN model.
Introducing global boxes as well as semantic boxes allows us to represent the page's semantic dependency structure more accurately.

Given a single-page image from a rich document, our semantic document layout analysis task is to predict a set of such bounding boxes together with their box types.
This setup follows classical layout analysis and can be formulated as a multi-class object detection problem.
Accordingly, we fine-tune pre-trained object detection models on a dataset with semantic layout labels.
In the rest of this section, we first describe how we construct the dataset used to train SCAN (Section~\ref{ssec:dataset-construction}), then explain how we fine-tune object detection models (Section~\ref{ssec:fine-tuning}), and finally detail the post-processing applied when integrating SCAN into textual and visual RAG pipelines (Section~\ref{ssec:post-processing}).

\begin{table}[t]
  \small
  \centering
  \scalebox{0.95}{
    \begin{tabular}{l|rr}
      \toprule
      \multicolumn{1}{c|}{\textbf{Cluster Domain}} & \multicolumn{1}{c}{\textbf{\# Pages}} & \multicolumn{1}{c}{\textbf{\# Boxes}} \\
      \midrule
      Tables, Charts                               & 3,766                                 & 17,024                                \\
      Flyers, Magazines, Menus, Recipes            & 3,747                                 & 21,555                                \\
      Maps, Travel information                  & 3,383                                 & 16,558                                \\
      Itemized documents                           & 3,771                                 & 19,062                                \\
      Handwritten text                             & 821                                   & 3,481                                 \\
      Vertical text                               & 3,768                                 & 16,520                                \\
      Math                                         & 1,213                                 & 5,479                                 \\
      Manuals, Guidelines, Blueprints              & 4,108                                 & 23,472                                \\
      \bottomrule
    \end{tabular}
  }
  \caption{Statistics of document domains in our dataset.}
  \label{tab:docsemnet-stats}
\end{table}

\begin{table}[t]
  \small
  \centering
  \scalebox{0.95}{
    \begin{tabular}{l|rr}
      \toprule
      \multicolumn{1}{c|}{\textbf{Box Type}} & \multicolumn{1}{c}{\textbf{Avg. Boxes Per Page}} & \multicolumn{1}{c}{\textbf{\# Boxes}} \\
      \midrule
      semantic box                           & 3.13                                             & 76,862                                \\
      title                                  & 0.49                                             & 12,137                                \\
      header                                 & 0.38                                             & 9,416                                 \\
      footer                                 & 0.73                                             & 17,969                                \\
      date                                   & 0.12                                             & 2,888                                 \\
      author                                 & 0.16                                             & 3,879                                 \\
      All types                              & 5.01                                             & 123,151                               \\
      \bottomrule
    \end{tabular}
  }
  \caption{Statistics of box types in our dataset.}
  \label{tab:docsemnet-box-stats}
\end{table}

\subsection{Dataset Construction}\label{ssec:dataset-construction}

As we are the first to define semantic document layout analysis in this context, we developed our training and evaluation dataset through a rigorous annotation process.
We first collected document pages from the Japanese portion of the CCpdf corpus~\cite{ccpdf}.
To ensure diversity in our training data, we first performed agglomerative clustering on a small subset of the corpus using image embeddings with the MiniCPM Visual Embedding~\cite{minicpmv} model.
We then selected a balanced number of sample pages close to the centroid of each cluster and included them in our dataset; the statistics for each cluster are provided in Table~\ref{tab:docsemnet-stats}.
We finally labeled 24,577 pages and split them into 22,577 pages for training, 1,000 for validation, and 1,000 for testing.
Appendix~\ref{appendix:docsemnet-examples} provides examples of our dataset.

The annotation task required annotators to identify semantic chunks within each page and draw bounding boxes around them, followed by assigning a box type.
The box types we defined are \textit{semantic box}, \textit{title}, \textit{header}, \textit{footer}, \textit{date}, and \textit{author}.
Table~\ref{tab:docsemnet-box-stats} reports the number of boxes for each class and the average number of boxes per page.

We collaborated with a specialized data annotation company in Tokyo, engaging six expert Japanese annotators under formal contracts with reasonable payment.
We first conducted a pilot annotation on a small number of samples to establish a detailed guideline regarding box granularity, segmentation criteria, and the handling of ambiguous cases.
During the pilot stage, we measured inter-annotator agreement using a matching-based IoU (Intersection over Union) metric (described in Section~\ref{ssec:fine-tuning}) and obtained a score above 0.7.
Because the IoU metric is relatively sensitive to box misalignment, we considered the agreement sufficiently high.
Thus, we adopted a single-annotator-per-sample protocol for the main annotation phase.
We also regularly checked annotation quality and updated the guideline.
Appendix~\ref{appendix:annotation-guideline} provides details of our annotation guideline.

\begin{table}[t]
  \small
  \centering
  \scalebox{0.9}{
    \begin{tabular}{cc|cc}
      \toprule
      \textbf{Model}             & \textbf{Confidence} & \textbf{IoU (\%)} & \textbf{Coverage (\%)} \\ 
      \midrule
      \multirow{4}{*}{YOLO11-X}  & 0.2                 & 53.7              & \textbf{81.0}          \\  
                                 & 0.3                 & \textbf{58.0}     & 79.0                   \\  
                                 & 0.4                 & 55.8              & 73.1                   \\  
                                 & 0.5                 & 49.7              & 64.0                   \\  
      \midrule
      \multirow{4}{*}{RT-DETR-X} & 0.2                 & 48.9              & 78.0                   \\  
                                 & 0.3                 & 54.8              & 78.8                   \\  
                                 & 0.4                 & 58.4              & \textbf{79.0}          \\  
                                 & 0.5                 & \textbf{59.6}     & 77.7                   \\  
      \bottomrule
    \end{tabular}
  }
  \caption{The Intersection over Union (IoU) scores and coverage ratios of fine-tuned object detection models.
    We varied the confidence threshold for predicted bounding boxes from 0.2 to 0.5.}
  \label{tab:smc-model-iou}
\end{table}

\begin{table*}[t]
  \small
  \centering
  \scalebox{0.9}{
    \begin{tabular}{l|cccccc}
      \toprule
      \multicolumn{1}{c|}{\textbf{Dataset}} & \textbf{Domain}                                              & \textbf{Lang.} & \textbf{\# Docs} & \textbf{\# Pages} & \textbf{\# QA} & \textbf{QA Type} \\
      \midrule
      OHR-Bench                             & \makecell[c]{Academic, Administration, Finance,                                                                                                          \\ Law, Manual, Newspaper, Textbook} & En             & 1,261           & 8,561            & 8,498         & \makecell[c]{Text, Table, Formula,\\ Chart, Reading Order}  \\
      \midrule
      Allganize                             & \makecell[c]{Finance, Information Technology, Manufacturing,                                                                                             \\ Public Sector, Retail/Distribution} & Ja             & 64              & 2,167            & 300           & Text, Chart, Table    \\
      \midrule
      BizMMRAG                              & \makecell[c]{Finance, Government, Medical,                                                                                                               \\ Consulting Sectors, Wikipedia}                  & Ja             & 42              & 3,924            & 146           & Text, Chart, Table    \\
      \bottomrule
    \end{tabular}
  }
  \caption{Statistics of our evaluation datasets.}
  \label{tab:datasets}
\end{table*}

\subsection{Object Detection Model Fine-tuning}\label{ssec:fine-tuning}

There are two popular families of object detection architectures: the CNN-based (YOLO series~\cite{yolo1,yolo11}) and the Transformer-based (DETR series~\cite{detr-eccv-2020,rtdetr}).
We fine-tuned YOLO11-X\footnote{\url{https://docs.ultralytics.com/models/yolo11/}} and RT-DETR-X\footnote{\url{https://docs.ultralytics.com/models/rtdetr/}} supported by the Ultralytics framework (under the AGPL-3.0 license) to develop our SCAN models.

To evaluate the fine-tuned models in terms of bounding box granularity and precision,
we developed a matching-based IoU (Intersection over Union) metric.
This metric first uses the Hungarian algorithm to perform bipartite matching between predicted and ground-truth bounding boxes.
Then, it calculates an average IoU over the matched bounding box pairs and unmatched bounding boxes.
For unmatched bounding boxes, we assigned an IoU of 0 to penalize excessive or insufficient predictions.

Table~\ref{tab:smc-model-iou} shows the two models' IoU scores on the validation set of our dataset.
Because each predicted bounding box has its confidence score, we varied the confidence threshold to select the
optimal predicted bounding box set.
We also computed the coverage score of the selected bounding boxes, defined as the ratio of the area covered by the selected boxes to the total area of all ground-truth boxes.
With the optimal confidence threshold, the IoU score of the RT-DETR-X fine-tuned model is better than that of the YOLO11-X fine-tuned model.
RT-DETR-X (confidence: 0.5) achieved the best IoU score of 59.6.
However, its coverage score is the lowest (77.7) among all the RT-DETR-X settings.
Thus, we selected the model based on RT-DETR-X with a confidence threshold of 0.4 as our primary SCAN model.
Appendix~\ref{appendix:scan-output-examples} shows examples of predicted bounding boxes from our SCAN model.

\subsection{Post-processing for RAG}\label{ssec:post-processing}

We use the outputs of SCAN as a preprocessing step for both textual and visual RAG pipelines (Figure~\ref{fig:system-overview}).
Given a page image, SCAN predicts a set of semantic and global boxes, and we crop the corresponding sub-images from the original page.
For textual RAG, we convert each sub-image into text using a VLM-based OCR model.
We then concatenate the resulting texts into a single page-level sequence according to a reading order estimated from the box coordinates.
In our implementation, we use a simple rule-based reading order prediction: boxes are sorted by the $y$-coordinate of the upper-left corner and then by the $x$-coordinate.
The concatenated text is indexed and used as the document representation in downstream textual RAG systems.
For visual RAG, we directly use all predicted boxes (semantic and global) as individual image chunks.

\section{Experiments}
\label{sec:exp}
\subsection{Datasets and Settings}
We use three datasets to evaluate the RAG performance: one English dataset, OHR-Bench~\cite{ohr}, and two Japanese datasets, our in-house BizMMRAG and Allganize~\cite{allganize}.
Each dataset is used for both textual and visual RAG evaluation, which involves answering questions based on retrieved document content.
As far as we know, OHR-Bench is the first benchmark to systematically evaluate the cascading impact of OCR on RAG systems.
It enables step-by-step evaluation across OCR, retrieval, generation, and overall performance.
Table~\ref{tab:datasets} gives a detailed summary of these datasets.
Section~\ref{sssec:textual-rag-setting}, Section~\ref{sssec:visual-rag-setting}, and Appendix~\ref{appendix:experiment-details} provide more details on the experimental settings.

\begin{table*}[t]
  \small
  \centering
  \scalebox{0.9}{
    \begin{tabular}{l|cc}
      \toprule
      \multicolumn{1}{c|}{\textbf{Model}} & \textbf{Architecture} & \textbf{Training Data}            \\
      \midrule
      DiT                                 & DiT (304M)            & IIT-CDIP (42M) + PubLayNet (360K) \\
      DocLayout-YOLO                      & YOLO11-X (57M)        & DocLayNet (80k)                   \\
      Beehive                             & RT-DETR-L (43M)       & DocLayNet (80k)                   \\
      SCAN                                & RT-DETR-X (67M)       & Our Dataset (22k)                 \\
      \bottomrule
    \end{tabular}
  }
  \caption{Comparison of layout analysis models.}
  \label{tab:model-comparison}
\end{table*}

\subsubsection{Textual RAG Setting}\label{sssec:textual-rag-setting}
When evaluating English textual RAG performance, we use OHR-Bench~\cite{ohr} and follow the same evaluation protocol.
We use BGE-m3~\cite{bgem3} and BM25 as retrieval models, and \texttt{meta-llama/Llama-3.1-8B-Instruct} and \texttt{Qwen/Qwen2-7B-Instruct} as answer models to generate answers according to the retrieved top-2 results.
We use three metrics: (a) retrieval, which calculates LCS (Longest Common Subsequence) to measure evidence inclusion in retrieved content; (b) generation, which measures the F1-score of QA when provided with the ground truth page; and (c) overall, which calculates the F1-score of QA for the end-to-end RAG pipeline.
The F1-score is calculated using precision and recall of common tokens between the generated result and the ground truth.
The final scores are the average across the four combinations of two retrieval models and two answer models.

For the Japanese datasets BizMMRAG and Allganize, we employ \texttt{intfloat/\allowbreak multilingual-e5-large} as a retrieval model.
For each query, the top-5 retrieved results are fed to the answer generation model, which is GPT-4o (\texttt{gpt-4o-2024-08-06}).
To evaluate the generated answers, we adopt the LLM-as-a-judge framework~\cite{llmasajudge}, using GPT-4o to assign an integer score from 1 to 5 to each generated answer.
Answers receiving a score greater than 4 are considered correct (assigned a value of 1), while others are considered incorrect (assigned a value of 0).
Final accuracy is computed based on these binary scores.

To demonstrate the effectiveness of our SCAN method, we compare several text conversion methods:
(1) using VLMs directly for text conversion without any layout analysis;
(2) using fine-grained layout analysis methods, including DiT~\cite{li2022dit}, DocLayout-YOLO~\cite{doclayoutyolo}, and Beehive~\cite{docling}, followed by a VLM for text conversion;
and (3) our SCAN method followed by a VLM for text conversion.
For the VLMs, we use an OCR-specialized model, GOT~\cite{got}, as well as a general VLM, Qwen2.5-VL~\cite{Qwen2.5-VL}.
We also include three other models, Nougat~\cite{nougat}, olmOCR~\cite{olmocr}, and InternVL2.5~\cite{internvl} for the setting (1) as baselines.
For the layout analysis models, we list their architectures and training data in Table~\ref{tab:model-comparison}.
Note that our SCAN model's training data is much smaller than the other three layout analysis models.
For the settings (2) and (3), we apply the post-processing described in Section~\ref{ssec:post-processing}.

\begin{table*}[t]
  \small
  \centering
  \resizebox{\textwidth}{!}{
    \begin{tabular}{l|ccccc|c|ccccc|c|ccccc|c}
      \toprule
                                          & \multicolumn{6}{c|}{Retrieval} & \multicolumn{6}{c|}{Generation} & \multicolumn{6}{c}{Overall}                                                                                                                                                               \\
                                          & TXT                            & TAB                             & FOR                         & CHA  & RO  & ALL                    & TXT  & TAB  & FOR  & CHA  & RO   & ALL                    & TXT  & TAB  & FOR  & CHA  & RO   & ALL                    \\
      \midrule
      Ground Truth                        & 81.2                           & 69.6                            & 74.8                        & 70.3 & 9.8 & 70.0                   & 49.4 & 46.0 & 34.0 & 47.0 & 28.2 & 43.9                   & 45.0 & 34.6 & 28.0 & 32.9 & 18.7 & 36.1                   \\
      \midrule
      \multicolumn{19}{l}{\textit{OCR-specialized small VLM}}                                                                                                                                                                                                                                            \\
      \midrule
      Nougat-350M                         & 59.1                           & 32.7                            & 44.2                        & 11.3 & 4.4 & 40.9                   & 36.7 & 22.9 & 22.9 & 6.4  & 6.9  & 25.5                   & 33.5 & 18.4 & 19.4 & 5.8  & 3.6  & 14.5                   \\
      GOT-580M                            & 62.1                           & 41.0                            & 48.7                        & 17.4 & 3.7 & 45.4                   & 37.5 & 28.5 & 24.1 & 8.5  & 7.1  & 27.8                   & 35.3 & 22.9 & 20.1 & 8.2  & 5.3  & 24.6                   \\
      DiT-GOT-580M                        & 67.5                           & 50.3                            & 47.7                        & 35.4 & 4.6 & 51.9 ($+$6.5)          & 46.4 & 35.0 & 25.5 & 19.6 & 14.8 & 35.0 ($+$7.2)          & 41.9 & 26.8 & 21.3 & 15.4 & 10.4 & 29.6 ($+$5.0)          \\
      DocLayout-YOLO-GOT-580M             & 60.4                           & 45.5                            & 43.2                        & 32.8 & 4.4 & 46.8 ($+$1.4)          & 41.9 & 31.4 & 24.5 & 22.2 & 18.3 & 32.7 ($+$4.9)          & 38.5 & 23.9 & 20.2 & 16.1 & 12.1 & 27.5 ($+$2.9)          \\
      Beehive-GOT-580M                    & 65.2                           & 49.5                            & 48.7                        & 39.0 & 4.7 & 51.2 ($+$5.8)          & 45.7 & 33.3 & 26.4 & 23.7 & 28.1 & 35.9 ($+$8.1)          & 41.4 & 25.0 & 21.5 & 16.6 & 17.0 & 29.6 ($+$5.0)          \\
      \textbf{SCAN-GOT-580M (ours)}       & 68.5                           & 54.3                            & 50.7                        & 36.6 & 5.3 & \textbf{53.9} ($+$8.5) & 46.2 & 37.9 & 27.5 & 20.8 & 24.6 & \textbf{36.9} ($+$9.1) & 41.9 & 28.4 & 22.5 & 16.0 & 17.3 & \textbf{30.8} ($+$6.2) \\
      \midrule
      \multicolumn{19}{l}{\textit{VLM for OCR}}                                                                                                                                                                                                                                                          \\
      \midrule
      InternVL2.5-78B                     & 68.6                           & 57.9                            & 55.6                        & 45.1 & 2.7 & 56.2                   & 41.7 & 41.8 & 29.0 & 33.6 & 3.3  & 35.8                   & 38.2 & 31.0 & 23.3 & 22.9 & 3.1  & 29.6                   \\
      olmOCR-7B                           & 72.5                           & 58.4                            & 55.4                        & 24.8 & 5.0 & 56.6                   & 44.8 & 40.5 & 30.4 & 19.0 & 8.4  & 36.0                   & 40.6 & 30.3 & 23.7 & 12.8 & 7.1  & 29.6                   \\
      Qwen2.5-VL-72B                      & 75.1                           & 60.0                            & 60.0                        & 38.2 & 5.3 & 59.6                   & 44.3 & 42.1 & 31.8 & 27.0 & 11.6 & 37.5                   & 40.6 & 31.1 & 26.1 & 19.0 & 8.8  & 31.1                   \\
      DiT-Qwen2.5-VL-72B                  & 76.9                           & 57.7                            & 55.6                        & 44.6 & 5.4 & \textbf{59.7} ($+$0.1) & 48.7 & 41.8 & 29.7 & 26.6 & 24.3 & 39.8 ($+$2.3)          & 44.8 & 32.0 & 24.5 & 20.3 & 16.0 & 33.5 ($+$2.4)          \\
      DocLayout-YOLO-Qwen2.5-VL-72B       & 63.5                           & 12.4                            & 36.0                        & 11.7 & 5.4 & 36.3 ($-$23.3)         & 41.2 & 10.2 & 19.0 & 7.2  & 18.2 & 24.4 ($-$13.1)         & 38.4 & 10.0 & 16.3 & 7.7  & 12.5 & 22.4 ($-$8.7)          \\
      Beehive-Qwen2.5-VL-72B              & 73.3                           & 16.0                            & 42.3                        & 16.3 & 6.1 & 42.6 ($-$17.0)         & 46.6 & 11.3 & 21.3 & 10.2 & 27.2 & 28.2 ($-$9.3)          & 43.0 & 10.9 & 19.5 & 8.6  & 16.6 & 25.3 ($-$5.8)          \\
      \textbf{SCAN-Qwen2.5-VL-72B (ours)} & 75.7                           & 56.6                            & 57.3                        & 40.6 & 6.5 & 58.9 ($-$0.7)          & 48.4 & 43.3 & 31.9 & 27.6 & 26.7 & \textbf{40.8} ($+$3.3) & 44.4 & 31.9 & 26.6 & 20.6 & 17.7 & \textbf{33.8} ($+$2.7) \\
      \bottomrule
    \end{tabular}
  }
  \caption{
    Textual RAG results on OHR-Bench.
    Comparison of various OCR methods across different evaluation metrics.
    \textit{TXT}, \textit{TAB}, \textit{FOR}, \textit{CHA}, \textit{RO}, and \textit{ALL} represent text, table, formula, chart, reading order, and their average, respectively.
    The RO category includes questions that require identifying the correct reading order to associate information from separate paragraphs.
    \textit{Ground Truth} indicates the performance when using the ground truth page text for retrieval and generation.
    The bold values indicate the best performance in each category.
  }
  \label{tab:textual-rag-results-en}
\end{table*}

\subsubsection{Visual RAG Setting}\label{sssec:visual-rag-setting}
We also apply OHR-Bench, BizMMRAG, and Allganize to evaluate visual RAG performance.
We use ColQwen2-v1.0~\cite{colpali} as an image retrieval model with top-5 retrieval and \texttt{Qwen/Qwen2.5-VL-7B} as an answer model.

Similar to the textual RAG setting, we compare three chunking methods:
(1) using single-page images as a visual segment for retrieval and generation;
(2) using layout analysis methods, DiT, DocLayout-YOLO, and Beehive, to chunk page images into layout-based chunks;
and (3) our SCAN method to chunk page images into semantic chunks.

\begin{table*}[t]
  \small
  \centering
  \scalebox{0.9}{
    \begin{tabular}{l|ccc|c|ccc|c}
      \toprule
                                          & \multicolumn{4}{c|}{BizMMRAG} & \multicolumn{4}{c}{Allganize}                                                                               \\
                                          & TXT                           & CHA                           & TAB  & ALL                    & TXT  & CHA  & TAB  & ALL                    \\
      \midrule
      Qwen2.5-VL-72B                      & 85.0                          & 52.3                          & 69.1 & 68.8                   & 84.5 & 68.4 & 62.2 & 71.7                   \\
      DiT-Qwen2.5-VL-72B                  & 75.0                          & 54.6                          & 50.0 & 59.9 ($-$8.9)          & 90.1 & 67.1 & 62.2 & 73.2 ($+$1.5)          \\
      DocLayout-YOLO-Qwen2.5-VL-72B       & 61.7                          & 25.0                          & 28.6 & 38.4 ($-$30.4)         & 49.3 & 21.1 & 23.2 & 31.2 ($-$40.5)         \\
      Beehive-Qwen2.5-VL-72B              & 70.0                          & 29.6                          & 21.4 & 40.3 ($-$28.5)         & 61.3 & 40.8 & 26.8 & 43.0 ($-$28.7)         \\
      \textbf{SCAN-Qwen2.5-VL-72B (ours)} & 81.7                          & 72.7                          & 73.8 & \textbf{76.1} ($+$7.3) & 85.9 & 85.5 & 72.0 & \textbf{81.1} ($+$9.4) \\
      \bottomrule
    \end{tabular}
  }
  \caption{Textual RAG results on Japanese datasets: BizMMRAG and Allganize.}
  \label{tab:textual-rag-results-ja}
\end{table*}

\begin{table*}[t]
  \small
  \centering
  \scalebox{0.9}{
    \begin{tabular}{l|ccccc|c}
      \toprule
                           & TXT  & TAB  & FOR  & CHA  & RO   & ALL                    \\
      \midrule
      No chunking          & 84.0 & 68.6 & 71.5 & 58.7 & 67.9 & 70.2                   \\
      DiT                  & 80.7 & 63.8 & 62.2 & 51.9 & 66.0 & 64.9 ($-$5.3)          \\
      DocLayout-YOLO       & 72.2 & 57.9 & 58.3 & 47.5 & 62.4 & 59.6 ($-$10.6)         \\
      Beehive              & 73.2 & 60.1 & 64.2 & 43.8 & 87.6 & 65.8 ($-$4.4)          \\
      \textbf{SCAN (ours)} & 86.0 & 70.0 & 73.5 & 63.4 & 86.3 & \textbf{75.8} ($+$5.6) \\
      \bottomrule
    \end{tabular}
  }
  \caption{Visual RAG results on OHR-Bench.}
  \label{tab:visual-rag-results-en}
\end{table*}

\begin{table*}[t]
  \small
  \centering
  \scalebox{0.9}{
    \begin{tabular}{l|ccc|c|ccc|c}
      \toprule
                           & \multicolumn{4}{c|}{BizMMRAG} & \multicolumn{4}{c}{Allganize}                                                                                \\
                           & TXT                           & CHA                           & TAB  & ALL                     & TXT  & CHA  & TAB  & ALL                    \\
      \midrule
      No chunking          & 71.7                          & 56.8                          & 57.1 & 58.9                    & 75.9 & 71.1 & 62.5 & 69.9                   \\
      DiT                  & 61.7                          & 59.5                          & 63.6 & 61.6 ($+$2.7)           & 81.8 & 68.8 & 64.1 & 71.6 ($+$1.7)          \\
      DocLayout-YOLO       & 55.0                          & 54.8                          & 61.4 & 57.0 ($-$1.9)           & 69.5 & 68.8 & 65.8 & 68.0 ($-$1.9)          \\
      Beehive              & 66.7                          & 45.5                          & 52.4 & 54.8 ($-$4.1)           & 66.0 & 52.6 & 60.0 & 59.5 ($-$10.4)         \\
      \textbf{SCAN (ours)} & 75.0                          & 61.4                          & 71.4 & \textbf{69.3} ($+$10.4) & 84.4 & 67.1 & 75.0 & \textbf{75.5} ($+$5.6) \\
      \bottomrule
    \end{tabular}
  }
  \caption{Visual RAG results on Japanese datasets: BizMMRAG and Allganize.}
  \label{tab:visual-rag-results-ja}
\end{table*}

\subsection{Textual RAG Evaluation Results}

\textbf{OHR-Bench.}
Table~\ref{tab:textual-rag-results-en} presents the comprehensive evaluation results of our SCAN method applied to various VLMs for text conversion in textual RAG.
Among conventional approaches, VLMs for OCR demonstrate superior performance, followed by OCR-specialized small VLMs.

Our SCAN model can improve the performance of VLM-based text conversions.
Despite the strong baseline performance of Qwen2.5-VL-72B, which achieves an impressive overall score of 31.1\% (the ground truth is 36.1\%), applying SCAN further improves the performance to 33.8\%.
The performance gains are larger when applying SCAN to OCR-specialized small VLMs.
With GOT, our SCAN's improvement is 6.2 points, enabling this smaller model to achieve competitive performance comparable to much larger VLMs.
This finding has important implications for deployment scenarios with computational constraints, suggesting that our semantic layout analysis approach can help bridge the efficiency-performance gap.
The generation results exhibit similar improvement patterns.
The results also indicate that SCAN's enhancements for structured content elements such as reading order (RO) and tables (TAB) become increasingly significant.
This suggests that the semantic segmentation approach is particularly valuable for preserving the relationships between elements that have spatial dependencies.

On the other hand, applying SCAN slightly degrades the retrieval performance when using Qwen2.5-VL-72B.
This is because retrieval is a relatively simple task within the RAG pipeline, primarily requiring the correct identification of keywords rather than comprehensive document understanding.
In contrast, the subsequent question-answering stage demands precise and complete conversion of document content into text, where SCAN's semantic layout analysis proves particularly advantageous.

We can also see that fine-grained document analysis methods, DocLayout-YOLO~\cite{doclayoutyolo} and Beehive~\cite{docling}, substantially degrade overall performance when used with Qwen2.5-VL-72B.
The degradations are particularly severe for structured content types such as tables and charts.
These conventional layout analysis methods typically segment documents into small atomic regions, which also break the structure of documents.
In contrast, DiT and our semantic box approach improve the strong baseline of Qwen2.5-VL-72B.
Although DiT is generally categorized as a conventional fine-grained document layout analysis method, its outputs include coarser segments compared to DocLayout-YOLO and Beehive, demonstrating that relatively coarser segments are more suitable for VLMs.
Our SCAN further optimizes this level of granularity: it preserves the integrity of semantically coherent regions, maintains their holistic structure while still providing the organizational benefits of layout analysis.
This preservation of semantic unity enables VLMs to process each region with full contextual awareness, resulting in more accurate text conversion and, ultimately, superior RAG performance.

It is notable that SCAN demonstrates high performance on English document benchmarks despite being trained exclusively on Japanese documents.
This suggests that for common layout patterns, the impact of language may be less significant than the effectiveness of the coarse-grained segmentation approach itself.

\textbf{BizMMRAG and Allganize.}
Table~\ref{tab:textual-rag-results-ja} presents the textual RAG performance for Japanese document datasets.
We include DiT, DocLayout-YOLO, and Beehive layout analysis models in our experiments, as the performance of layout analysis is not heavily dependent on language.
The results have the same trends as the English OHR-Bench evaluation, demonstrating that our SCAN methodology yields substantial improvements over a capable VLM.
Specifically, on the BizMMRAG dataset, our SCAN-enhanced approach demonstrates a notable 7.3-point improvement compared to the baseline Qwen2.5-VL-72B model: text accuracy decreased by 3.3 points, while chart accuracy increased by 20.4 points and table accuracy increased by 4.7 points.
We observe similar trends for the Allganize dataset.

We also observe that the performance improvements are larger for Japanese datasets than for the English OHR-Bench.
When using Qwen2.5-VL-72B, the gain was 2.7 points on OHR-Bench, but it increases to 7.3 points on BizMMRAG and 9.4 points on Allganize.
In general, multilingual VLMs tend to demonstrate relatively higher performance in major languages such as English.
Thus, while they can achieve reasonable performance on English documents even when the input images are complex, their performance is likely to degrade substantially on Japanese documents due to greater layout and linguistic complexity.
This indicates that SCAN, which mitigates the input image complexity for VLMs, is especially effective for non-major languages such as Japanese, yielding even greater benefits than in English.

\subsection{Visual RAG Evaluation Results}\label{sec:visual:result}
\textbf{OHR-Bench.}
Table~\ref{tab:visual-rag-results-en} presents the results of OHR-Bench in visual RAG.
When applying our SCAN approach to divide original pages into semantic chunks and performing visual RAG on these chunks, we observed an overall improvement of 5.6 points compared to processing entire pages.
We can see that the SCAN approach is effective for every category.
Especially in the RO (reading order) task, our method achieves an impressive 18.4-point improvement.
Recall that the RO task requires examining different paragraphs and articles to summarize answers.
This demonstrates that dividing a page image into independent semantic chunks enables the system to retrieve only the relevant paragraphs, avoiding distractions from unrelated content on the same page.

\textbf{BizMMRAG and Allganize.}
Table~\ref{tab:visual-rag-results-ja} presents the visual RAG results for Japanese document datasets.
The findings demonstrate that on both the BizMMRAG and Allganize benchmarks, our SCAN methodology exhibits substantial accuracy improvements.
Specifically, SCAN improves by 10.4 points on BizMMRAG and 5.6 points on Allganize.
This result also shows that our SCAN approach enables the VLM to achieve significantly enhanced performance in Japanese VQA.

\begin{table}[t]
  \small
  \centering
  \scalebox{0.9}{
    \begin{tabular}{l|rrrr}
      \toprule
      \multicolumn{1}{c|}{\textbf{Setting}} & \textbf{\makecell[c]{\# Input                         \\Tokens}} & \textbf{\makecell[c]{\# Output\\Tokens}} & \textbf{\# Chunks} & \textbf{Time (s)} \\
      \midrule
      No chunking                           & 1,320.4                       & 991.9   & 1.0  & 68.0 \\
      \textbf{SCAN}                         & 9,683.1                       & 2,515.0 & 12.4 & 56.3 \\
      \bottomrule
    \end{tabular}
  }
  \caption{
    Processing cost and time comparison of VLM text conversion.
    We used 10 images randomly sampled from OHR-Bench, and the values in the table are averages over these 10 instances.
  }
  \label{tab:cost}
\end{table}

\subsection{Cost Comparison of VLM Text Conversion}
\label{apd:sec:cost}
Our approach, which splits a page into multiple images, consistently improves the accuracy of textual RAG.
However, one concern is that increasing the number of images to be processed may also increase processing time and cost.
To investigate this, we compared the processing time and token usage of VLM-based text conversion with and without applying SCAN.

\begin{table*}[t]
  \small
  \centering
  \begin{tabular}{l|lrrrr}
    \toprule
    \multicolumn{1}{c|}{\textbf{Chunking Method}} & \textbf{\makecell[c]{Model Architecture                                                \\(\# Parameters)}} & \textbf{\makecell[c]{\# Chunks                                         \\Per Image}} & \textbf{\makecell[c]{Relative Chunk                                 \\Size (\%)}} & \textbf{\makecell[c]{Textual RAG\\Score}}   & \textbf{\makecell[c]{Visual RAG\\Score}}    \\
    \midrule
    No chunking                                   & N/A                                     & 1.0  & 100.0 & 31.1          & 70.2          \\
    DiT                                           & DiT (304M)                              & 12.3 & 16.3  & 33.5          & 64.9          \\
    DocLayout-YOLO                                & YOLO11-X (57M)                        & 9.9  & 11.3  & 22.4          & 59.6          \\
    Beehive                                       & RT-DETR-L (43M)                       & 17.4 & 4.8   & 25.3          & 65.8          \\
    \textbf{SCAN$_{\text{YOLO}}$}                 & YOLO11-X (57M)                        & 3.2  & 26.4  & 33.5          & 72.4          \\
    \textbf{SCAN$_{\text{RT-DETR}}$}              & RT-DETR-X (67M)                       & 5.2  & 19.1  & \textbf{33.8} & \textbf{75.8} \\
    \bottomrule
  \end{tabular}
  \caption{
    Effect of model architecture and chunking granularity on RAG performance.
    Textual and visual RAG scores are copied from Tables~\ref{tab:textual-rag-results-en} and \ref{tab:visual-rag-results-en}, respectively.
    SCAN$_{\text{YOLO}}$ is a model trained on the same data as SCAN$_{\text{RT-DETR}}$ but uses the YOLO11-X architecture.
    We used the same confidence threshold (0.4) as SCAN$_{\text{RT-DETR}}$ for SCAN$_{\text{YOLO}}$.
    Relative chunk size is computed as the average area of chunks divided by the area of the original page image.
    We randomly sampled 100 images from OHR-Bench for this evaluation.
  }
  \label{tab:granularity-and-rag-performance}
\end{table*}

We randomly sampled 10 images from OHR-Bench and used Qwen2.5-VL-72B for text conversion.\footnote{We used a vLLM \cite{vllm} server running on four NVIDIA RTX PRO 6000 Blackwell GPUs for efficient batch processing.}
Table~\ref{tab:cost} reports the comparison between single-page processing and multiple semantic chunk processing.
The results show that both the number of input tokens and output tokens increase when using multiple semantic chunks.
This implies that applying SCAN with API-based models that charge by token usage may lead to higher cost.
However, despite the increase in token counts, the average processing time is reduced.
We attribute this to the substantial decrease in the number of input tokens per request to the VLM, which lowers the cost of attention computation.
Specifically, while the average number of input tokens per chunk is 1,320.4 in the single-page setting, it is reduced to 780.9 ($= 9683.1 / 12.4$) in the multiple-chunk setting.
Thus, SCAN not only enables the extraction of richer textual information but also reduces computational overhead.

\subsection{Chunking Granularity and RAG Performance}
To quantitatively evaluate how chunking granularity affects RAG performance, we computed the average number of chunks per image and their relative sizes for the five different chunking methods (Table~\ref{tab:granularity-and-rag-performance}).
From the number and relative size of chunks, we observe that our SCAN divides a page into relatively larger and fewer chunks compared to the other chunking methods.
SCAN achieves the highest performance in both textual and visual RAG, indicating that its moderate level of granularity is well-suited for RAG tasks.

To ablate the effect of chunking granularity, we trained another SCAN model using the YOLO11-X backbone (denoted as SCAN$_{\text{YOLO}}$ in Table~\ref{tab:granularity-and-rag-performance}).
SCAN$_{\text{YOLO}}$ and SCAN$_{\text{RT-DETR}}$ perform comparably in textual and visual RAG, suggesting the model architecture has a minor impact on RAG performance.
Comparing SCAN$_{\text{YOLO}}$ with DocLayout-YOLO, despite sharing the same architecture as DocLayout-YOLO, SCAN$_{\text{YOLO}}$ yields substantially higher performance in both RAG settings, even though it is trained on significantly fewer annotations (23k vs. 80k).
These results demonstrate that the performance gains stem from our coarse granularity policy rather than model capacity or dataset scale.

\section{Conclusion}
We presented SCAN, a semantic document layout analysis approach for modern textual and visual RAG systems.
By introducing coarse-grained semantic segmentation that preserves topical coherence, SCAN effectively reduces the information processing burden on VLMs while maintaining semantic integrity across document components.
To develop SCAN, we labeled more than 24k document images with semantic layouts and trained a robust semantic layout analysis model.
Our comprehensive evaluation across multiple datasets, languages, and document types demonstrates SCAN's ability to enhance textual and visual RAG performance by 2.7--9.4 and 5.6--10.4 points, respectively.
In addition, SCAN achieves these improvements while reducing computational costs.


Future work could explore SCAN's applicability to other document understanding tasks,
such as document summarization, information extraction, and document VQA.

\section*{Ethical Statement}
In this work, we study semantic document layout analysis for RAG.
To the best of our knowledge, there is no negative societal impact in this research.
All our training data consist of publicly available PDFs from the internet, which likewise present no ethical concerns.
Our SCAN model aims to improve information extraction without introducing biases in the underlying content.
We believe that improved document analysis can enhance the accessibility of information for users across different languages and document formats.
While our system improves RAG capabilities, users should still be mindful of the general limitations of AI systems when relying on generated answers.

We used Claude-3.7-Sonnet and GPT-5 to help polish the writing of the paper.
We are responsible for all the materials presented in this work.

\section*{Limitations}
While our SCAN approach offers significant advantages, we acknowledge several limitations that present opportunities for future research:

1. Our SCAN model operates based on spatial image layout.
In certain documents where content that should logically form a single semantic chunk is physically separated in space and does not fit in a rectangular box, our current model cannot yet establish these connections.
This limitation could potentially be addressed through an additional trainable reading order model coupled with a semantic box merging mechanism.

2. Our current model was trained primarily on Japanese data.
While experiments demonstrate improvements on English benchmarks as well, this may not represent the optimal model for all languages.
Japanese documents have unique layout characteristics, such as vertical writing and right-to-left orientation, which differ from English conventions.
Further analysis and exploration are needed, and future work could involve annotating purely English data to investigate whether higher performance could be achieved for English RAG applications.

3. SCAN's semantic layout was designed for dense, content-rich document RAG.
For simpler pages, designing an adaptive approach that intelligently decides whether to apply semantic layout analysis or process the page as a single unit could provide better generalizability in future iterations.

\bibliography{custom.bib}

\clearpage
\onecolumn
\appendix

\section{Examples from Our Semantic Layout Dataset}\label{appendix:docsemnet-examples}
\begin{figure*}[htbp]
  \centering
  \begin{minipage}{0.45\textwidth}
    \centering
    \includegraphics[width=1\textwidth]{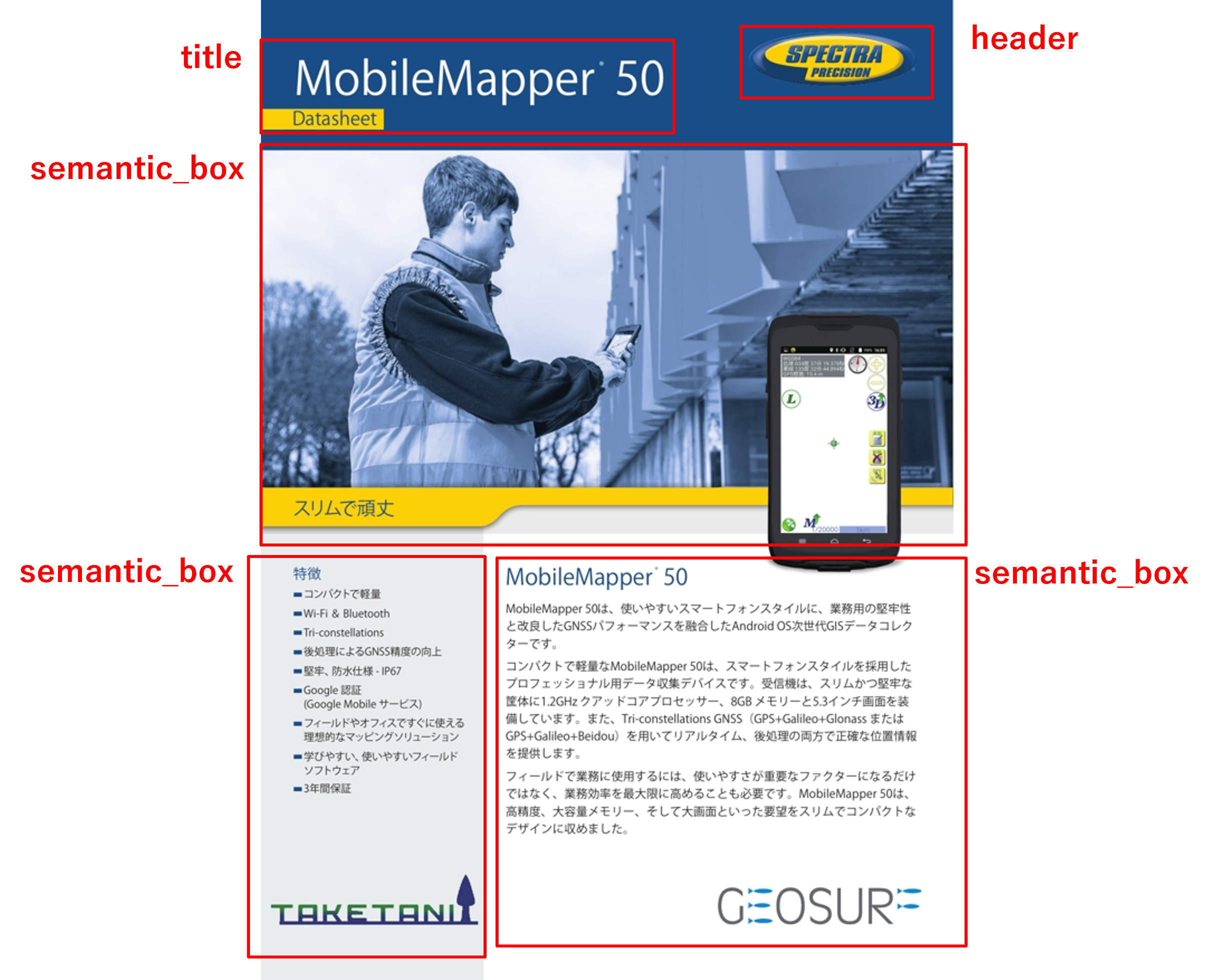}
  \end{minipage}
  \begin{minipage}{0.45\textwidth}
    \centering
    \includegraphics[width=1\textwidth]{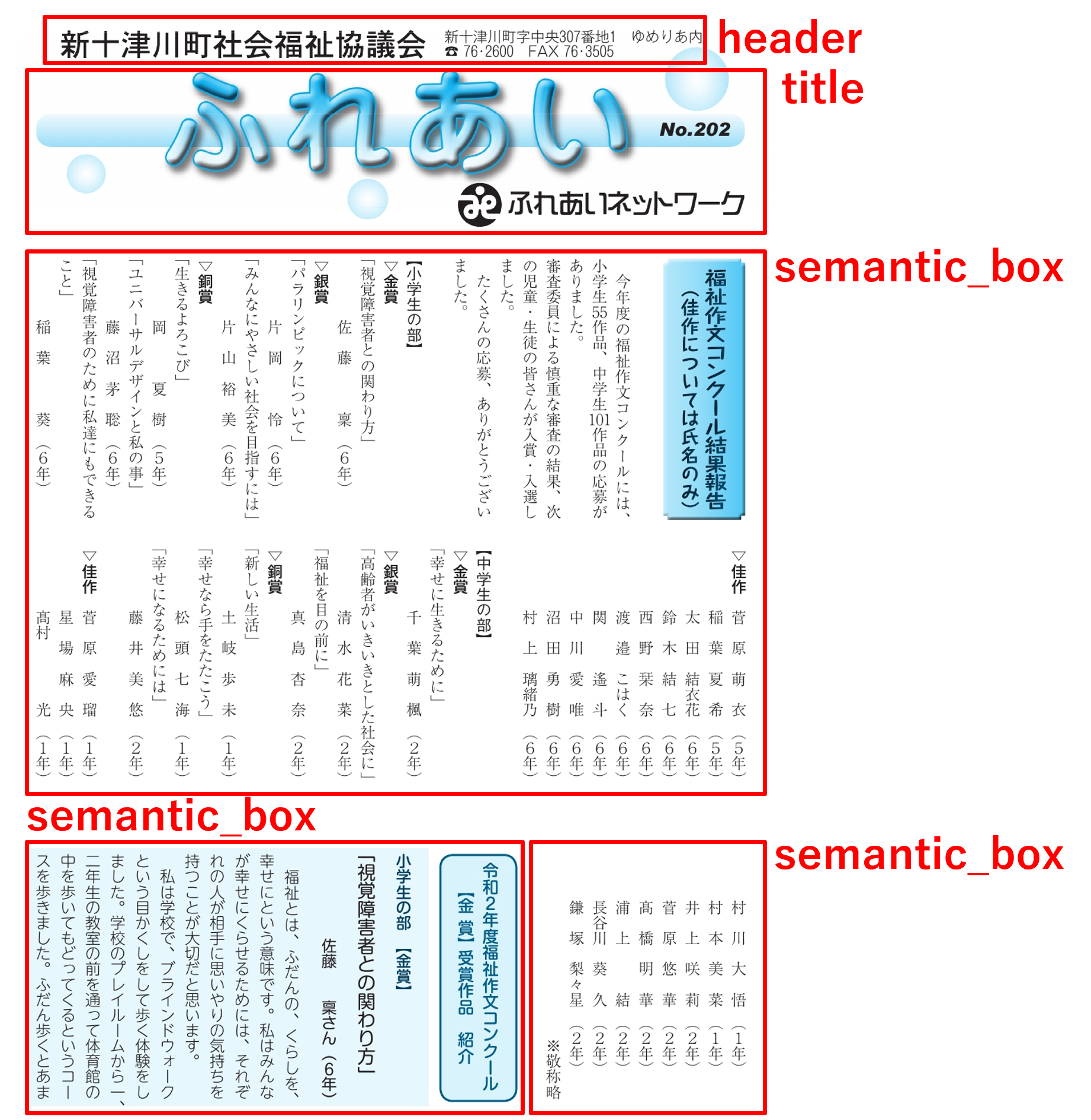}
  \end{minipage}
  \centering
  \begin{minipage}{0.45\textwidth}
    \centering
    \includegraphics[width=1\textwidth]{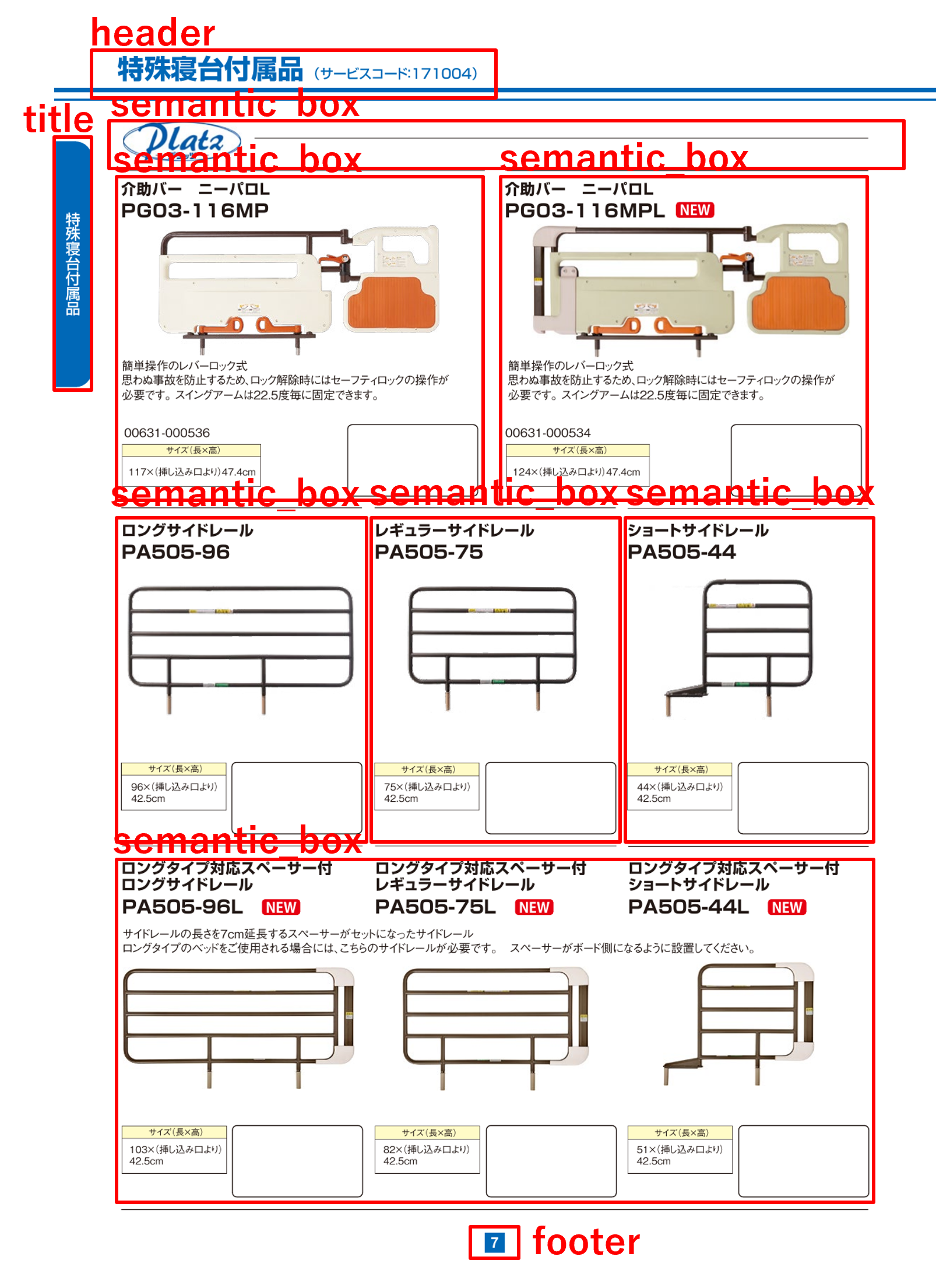}
  \end{minipage}
  \begin{minipage}{0.45\textwidth}
    \centering
    \includegraphics[width=1\textwidth]{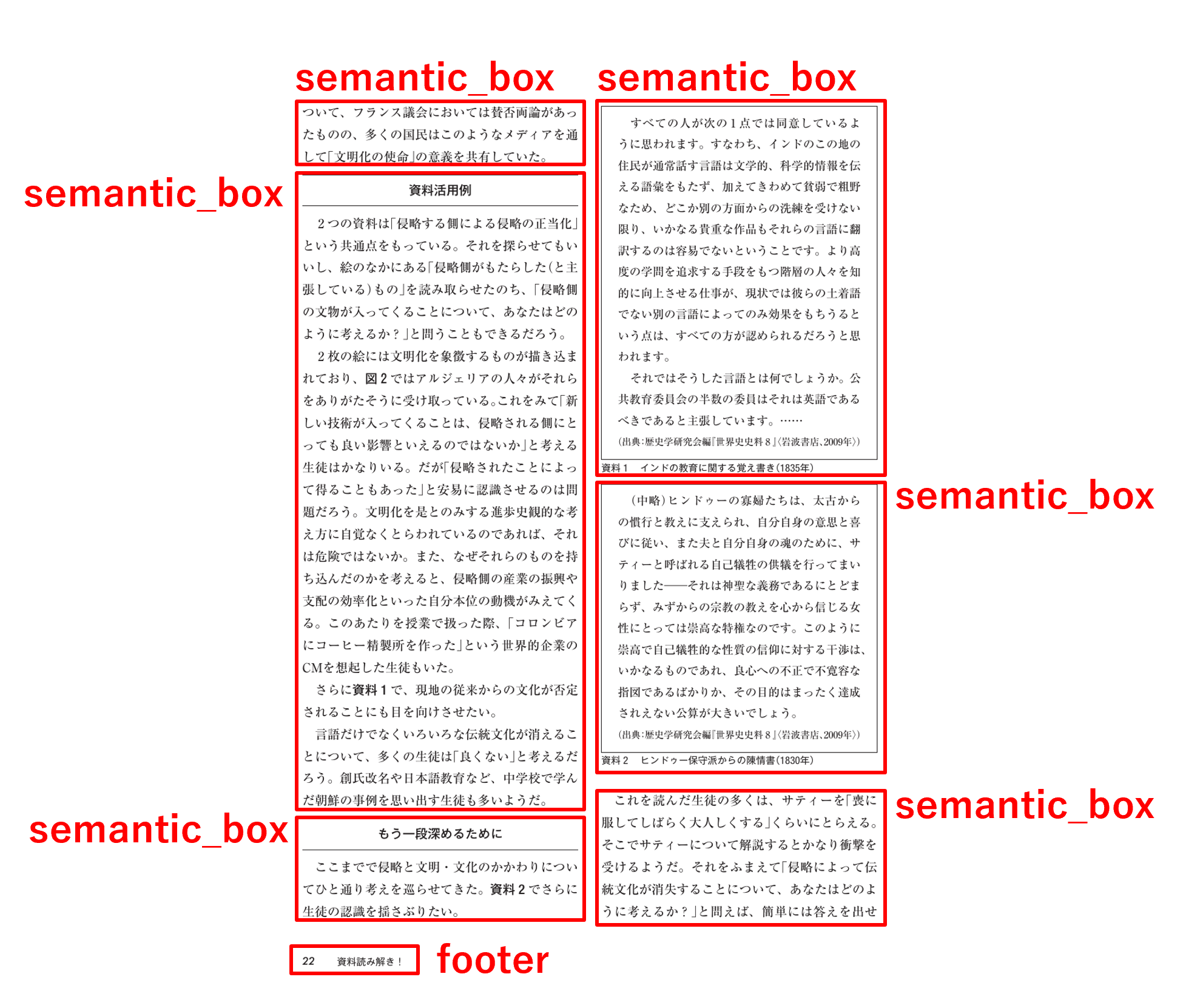}
  \end{minipage}
  \caption{Examples of our semantic layout dataset.}
  \label{fig:annotation}
\end{figure*}

Figure~\ref{fig:annotation} shows some examples of our semantic layout dataset.
It contains diverse document pages, including research papers, administrative reports, user manuals, slides, flyers, and more.
The dataset is annotated with the \textit{semantic box} class and the five global box classes: \textit{title}, \textit{header}, \textit{footer}, \textit{date}, and \textit{author}.

\section{Annotation Instructions for Our Semantic Layout Dataset}\label{appendix:annotation-guideline}
This section summarizes the guideline we provided to annotators when creating our semantic layout dataset.
Annotators first locate and annotate page-level global boxes such as headers and footers.
After that, they segment the remaining content into subtopics so that each subtopic is as semantically self-contained as possible.
For example, in a product catalog (e.g., the bottom-left example in Figure~\ref{fig:annotation}), the description of each product is independent of the others and therefore receives its own semantic box.
If a single subtopic still contains too much information, annotators further split it so that each semantic box contains a manageable amount of content, which is suitable for processing by VLMs.

Semantic units are always annotated with axis-aligned rectangles.
Rectangles are required not to overlap and, taken together, should cover all content on the page.
When a single rectangle would inevitably contain content from multiple topics, the region is split so that each rectangle corresponds to a single topic.
For instance, in the top-right example in Figure~\ref{fig:annotation}, the bottom-right semantic box is topically related to the upper box, but they are annotated separately because a single rectangle would also include the bottom-left region, which belongs to a different topic.

During annotation, we discard pages that are unsuitable for our task.
Typical exclusion cases include:
\begin{itemize}
  \item pages written in languages other than Japanese or with text so small that the content cannot be reliably read
  \item pages that are heavily rotated or not displayed in the correct orientation
\end{itemize}

\section{Training Details of Our Object Detection Models}
\label{appendix:training-details}

We fine-tuned two off-the-shelf object detection models: YOLO11-X and RT-DETR-X.
We mostly followed the default settings provided by the Ultralytics framework (version 8.3.28)\footnote{\url{https://github.com/ultralytics/ultralytics/releases/tag/v8.3.28}}, but we explicitly set or tuned some important hyper-parameters as shown in Table~\ref{tab:hyper-parameters}.
For the YOLO11-X fine-tuning, we used 4 NVIDIA L40 GPUs (48GB), which took about 4 hours to finish 30 epochs.
For the RT-DETR-X fine-tuning, we used 8 NVIDIA A100 GPUs (80GB), which took about 16 hours to finish 120 epochs.

\begin{table*}[h]
  \centering
  \small
  \begin{tabular}{l|cc}
    \toprule \textbf{Hyper-parameter} & \textbf{YOLO11-X}                   & \textbf{RT-DETR-X}                        \\
    \midrule Batch size               & \{8, 16, \textbf{32}, 64\}          & \{8, 16, 32, \textbf{64}\}                \\
    Learning rate                     & \{\textbf{1e-4}, 5e-4, 1e-3, 5e-3\} & \{5e-5, 1e-4, \textbf{5e-4}, 1e-3, 5e-3\} \\
    Max training epochs               & \{\textbf{30}, 40, 80, 120\}        & \{80, \textbf{120}, 160\}                 \\
    Weight decay                      & \{5e-5, 1e-4, \textbf{5e-4}\}       & \{1e-5, \textbf{1e-4}, 1e-3\}             \\
    Warmup epochs                     & \{\textbf{5}, 10\}                  & \{5, \textbf{10}\}                        \\
    Image size                        & \multicolumn{2}{c}{1024}                                                        \\
    Dropout                           & \multicolumn{2}{c}{0.0}                                                         \\
    Optimizer                         & \multicolumn{2}{c}{AdamW}                                                       \\
    Learning rate scheduler           & \multicolumn{2}{c}{cos\_lr}                                                     \\
    \bottomrule
  \end{tabular}
  \caption{Hyper-parameters used for fine-tuning object detection models. We tuned
    the hyper-parameters in the brackets in terms of the mean average precision (mAP)
    on the validation set. The bold values are the best hyper-parameters for each
    model.}
  \label{tab:hyper-parameters}
\end{table*}

\section{Output Examples of Our Semantic Layout Analysis Model}\label{appendix:scan-output-examples}

\begin{figure}[ht]
  \centering
  \includegraphics[width=0.9\textwidth]{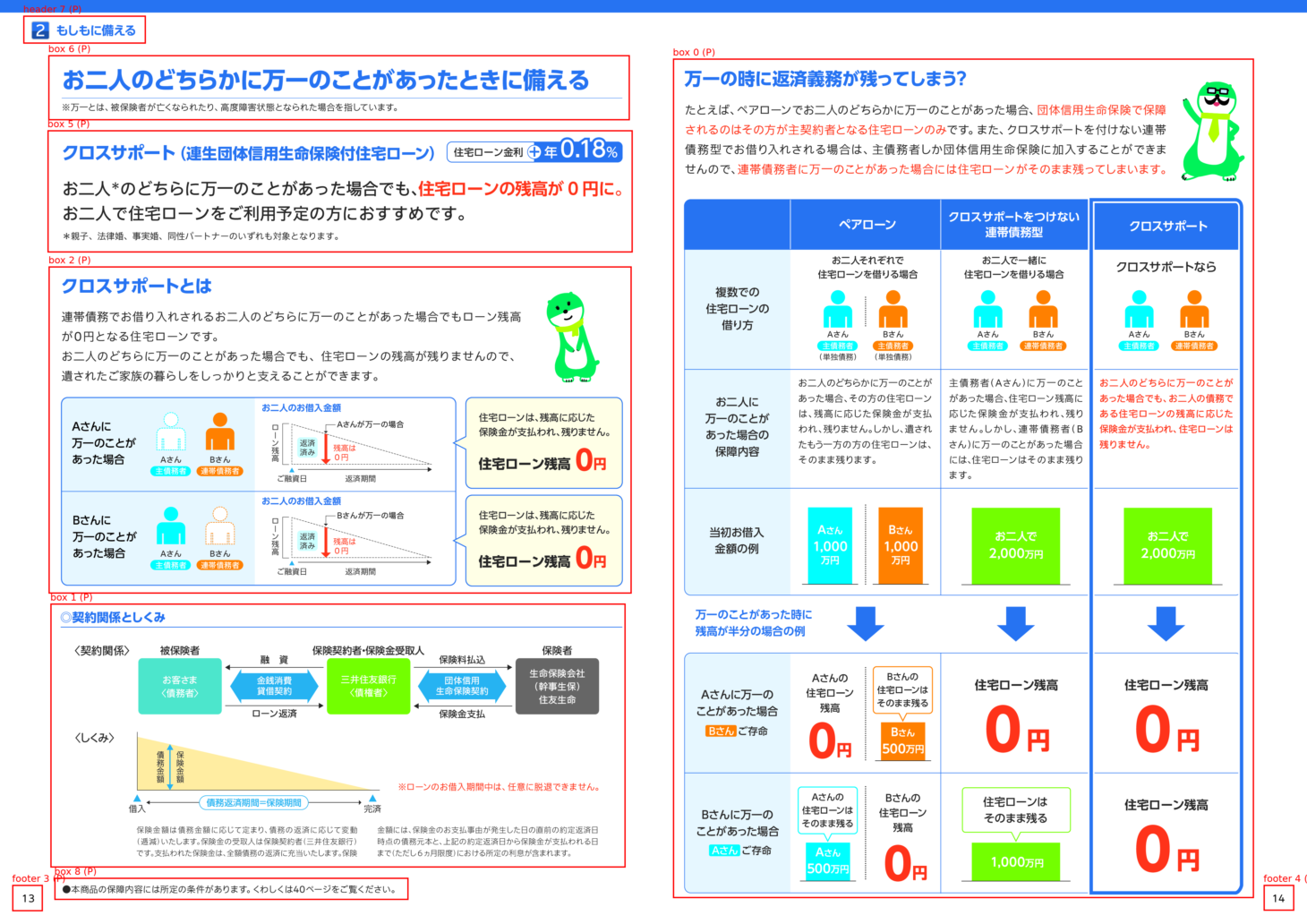}
  \caption{Examples of SCAN model outputs.}
  \label{fig:smc-result-1}
\end{figure}

\begin{figure}[ht]
  \centering
  \includegraphics[width=0.7\textwidth]{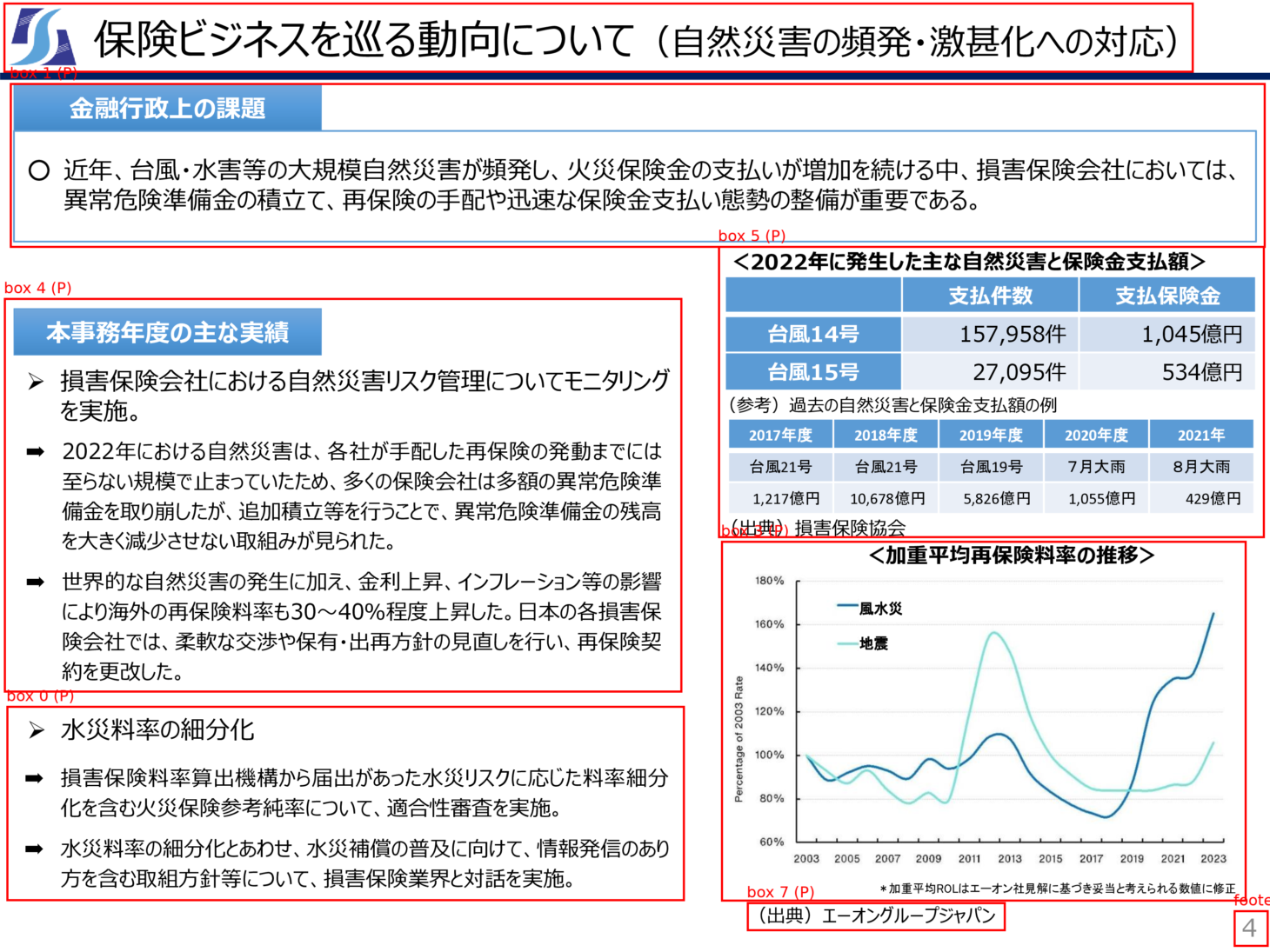}
  \caption{Example of SCAN model outputs.}
  \label{fig:smc-result-2}
\end{figure}

\begin{figure}[ht]
  \centering
  \includegraphics[width=0.75\textwidth]{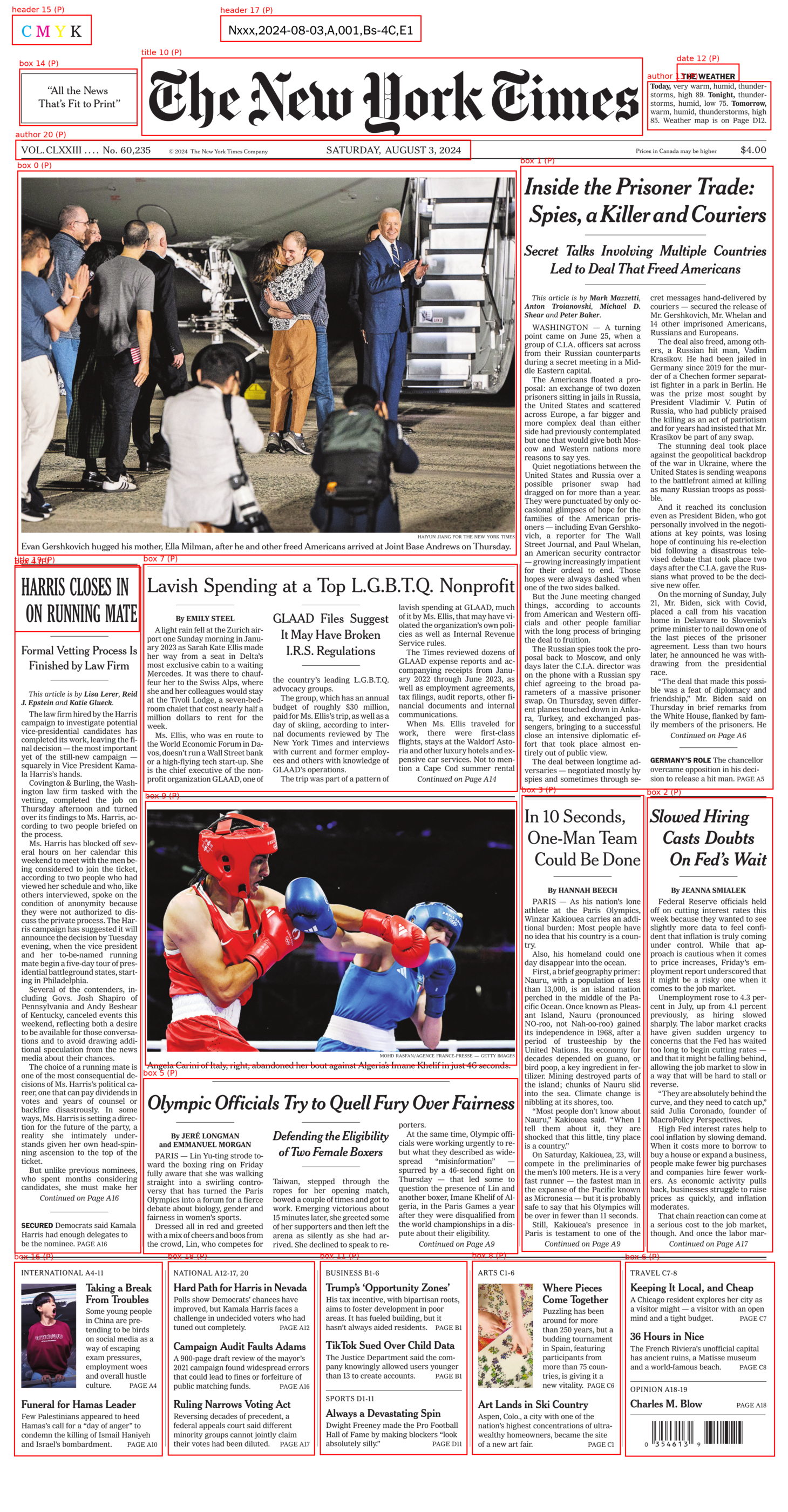}
  \caption{Example of SCAN model outputs.}
  \label{fig:smc-result-3}
\end{figure}

\begin{figure}[ht]
  \centering
  \includegraphics[width=0.9\textwidth]{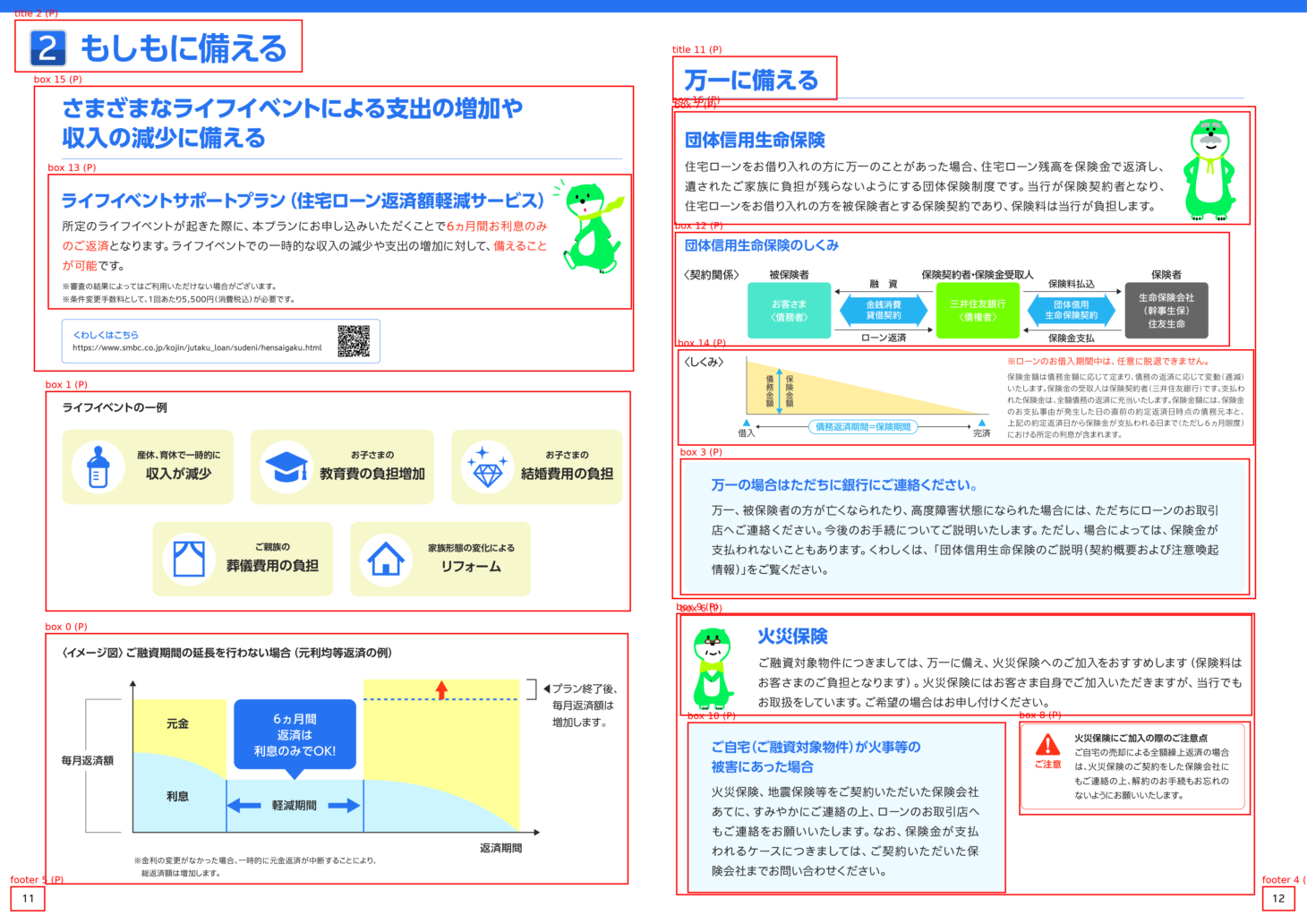}
  \caption{Example of SCAN model outputs.}
  \label{fig:smc-result-4}
\end{figure}

\begin{figure}[ht]
  \centering
  \includegraphics[width=0.8\textwidth]{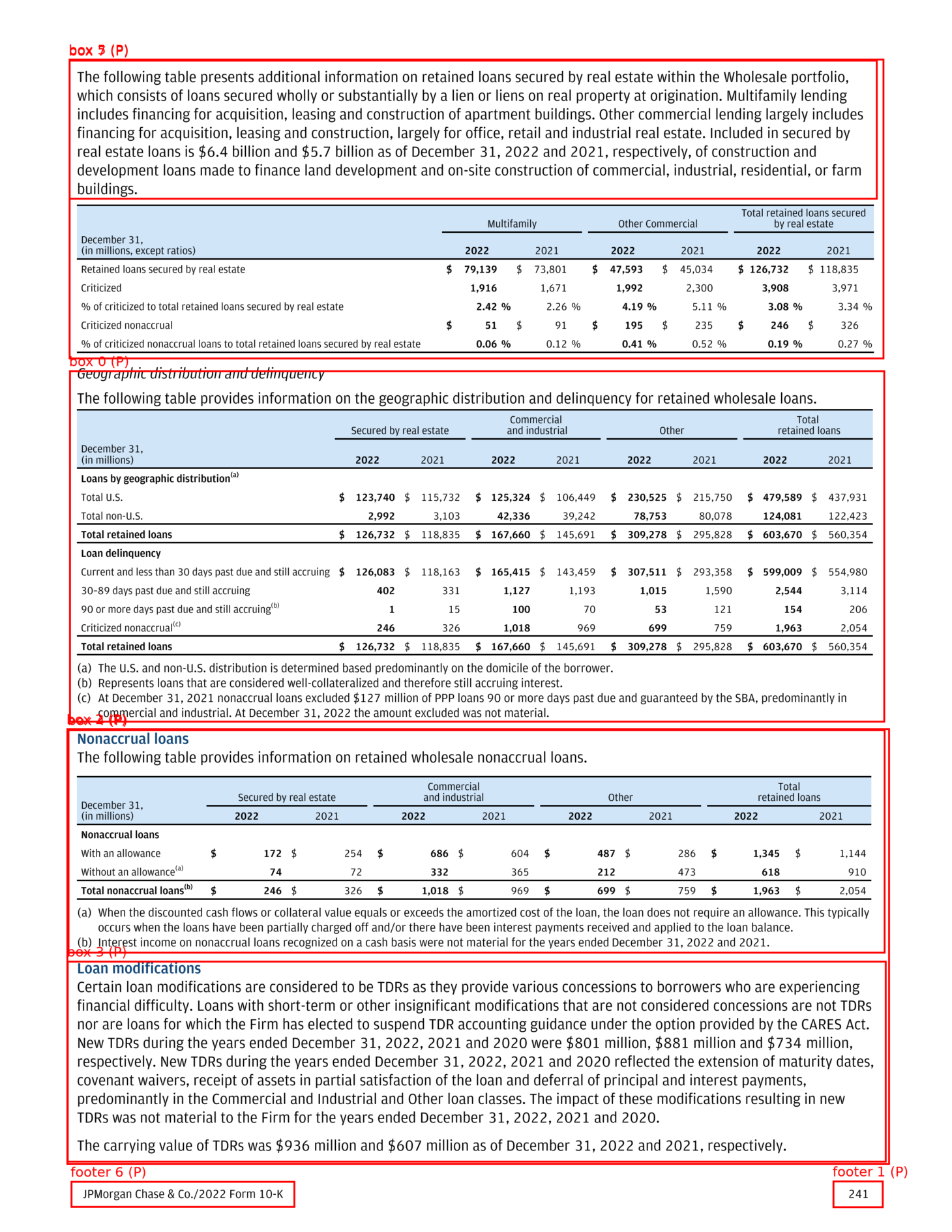}
  \caption{Example of SCAN model outputs.}
  \label{fig:smc-result-5}
\end{figure}

This section presents qualitative examples of SCAN predictions.
Figures~\ref{fig:smc-result-1}--\ref{fig:smc-result-5} illustrate that SCAN can handle complex layouts in both Japanese and English documents, grouping related elements into coherent semantic chunks.

Figures~\ref{fig:smc-result-4} and \ref{fig:smc-result-5} also reveal typical failure patterns, where some regions are covered by overlapping boxes.
Such redundancy is usually harmless when semantic chunks are used as retrieval units in RAG or as inputs to LLMs, but it may be undesirable in traditional OCR pipelines that aim to extract each character exactly once.
For these use cases, it would be beneficial to add post-processing that merges or removes overlapping boxes.

\clearpage

\section{Experimental Details}\label{appendix:experiment-details}

\subsection{Environment}
For text conversion with Qwen2.5-VL-72B, we used 8 NVIDIA L40S (48GB) GPUs and an INTEL(R) XEON(R) GOLD 6548N CPU. The details are as follows.
\begin{lstlisting}[frame=single]
  python 3.12
  vllm==0.7.3 (V0 version)
  torch==2.5.1
  torchaudio==2.5.1
  torchvision==0.20.1
  transformers==4.49.0
  ultralytics==8.3.28
  vLLM settings
  - temperature: 0.3
  - top_p: 0.95
  - max_tokens: 8192
  - repetition_penalty: 1.1
  - tensor_parallel==4
\end{lstlisting}

\subsection{Prompts for the VLM Text Conversion}
\begin{lstlisting}[frame=single]
You are a powerful OCR assistant tasked with converting PDF images to the Markdown format. You MUST obey the following criteria:
1. Plain text processing:
- Accurately recognize all text content in the PDF image without guessing or inferring.
- Precisely recognize all text in the PDF image without making assumptions in the Markdown format.
- Maintain the original document structure, including headings, paragraphs, lists, etc.
2. Formula Processing:
- Convert all formulas to LaTeX.
- Enclose inline formulas with $ $. For example: This is an inline formula $ E = mc^2 $.
- Enclose block formulas with $$ $$. For example: $$ \frac{-b \pm \sqrt{b^2 - 4ac}}{2a} $$.
3. Table Processing:
- Convert all tables to LaTeX format.
- Enclose the tabular data with \begin{table} \end{table}.
4. Chart Processing:
- Convert all Charts to LaTeX format.
- Enclose the chart data in tabular with \begin{table} \end{table}.
5. Figure Handling:
- Ignore figures from the PDF image; do not describe or convert images.
6. Output Format:
- Ensure the Markdown output has a clear structure with appropriate line breaks.
- Maintain the original layout and format as closely as possible.
Please strictly follow these guidelines to ensure accuracy and consistency in the conversion. Your task is to accurately convert the content of the PDF image using these format requirements without adding any extra explanations or comments.
\end{lstlisting}

\subsection{Prompts for LLM-as-a-judge}
\begin{lstlisting}[frame=single]
System:
You are an expert evaluation system for a question answering chatbot.

You are given the following information:
- a user query and reference answer
- a generated answer

You may also be given a reference answer to use for reference in your evaluation.

Your job is to judge the relevance and correctness of the generated answer.
Output a single score that represents a holistic evaluation.
You must return your response in a line with only the score.
Do not return answers in any other format.
On a separate line provide your reasoning for the score as well.

Follow these guidelines for scoring:
- Your score has to be between 1 and 5, where 1 is the worst and 5 is the best.
- Your output format should be in JSON with fields "reason" and "score" shown below.
- If the generated answer is not relevant to the user query, you should give a score of 1.
- If the generated answer is relevant but contains mistakes, you should give a score between 2 and 3.
- If the generated answer is relevant and fully correct, you should give a score between 4 and 5.

Example Response in JSON format:
{{
    "reason": "The generated answer has the exact same metrics as the reference answer, but it is not as concise.",
    "score": "4.0"

}}

User:
## User Query
{query}

## Reference Answer
{reference_answer}

## Generated Answer
{generated_answer}
\end{lstlisting}

\end{document}